\documentclass{article}
\usepackage{spconf,amsmath,graphicx}
\usepackage{algorithm}
\usepackage{algorithmic}
\usepackage{makecell}

\title{Judging from Support-set: A New Way to Utilize Few-Shot Segmentation for Segmentation Refinement Process}
\name{Seonghyeon Moon\textsuperscript{\rm 1,2}, Qingze (Tony) Liu\textsuperscript{\rm 3}, Haein Kong\textsuperscript{\rm 3}, Muhammad Haris Khan\textsuperscript{\rm 4}}
\address{    \textsuperscript{\rm 1}Brookhaven National Laboratory 
    \textsuperscript{\rm 2}Roblox\\
    \textsuperscript{\rm 3}Rutgers University 
    \textsuperscript{\rm 4}Mohamed Bin Zayed University of Artificial Intelligence}

\begin{document}
%
\maketitle
\begin{abstract}
Segmentation refinement enhances coarse masks generated by segmentation algorithms, aiming for detailed and accurate contours of target objects. Despite advancements in segmentation refinement research, no method exists to evaluate its success, which is critical for reliable applications. To address this gap, we propose Judging From Support-set (JFS), leveraging a few-shot segmentation (FSS) model in a novel evaluation pipeline. Traditional FSS aims to locate target objects in query images using support set information. In JFS, coarse and refined masks from segmentation refinement methods become support masks for the FSS model, with the existing support mask serving as the test set. This setup evaluates the quality of refined segmentation. We validate JFS using the SAM Enhanced Pseudo-Labels (SEPL) and SegGPT on the PASCAL dataset, demonstrating its potential to reliably judge segmentation refinement success and foster innovation in image processing.
\end{abstract}

\begin{figure}[ht!]
 \centering 
 \includegraphics[width=18em]{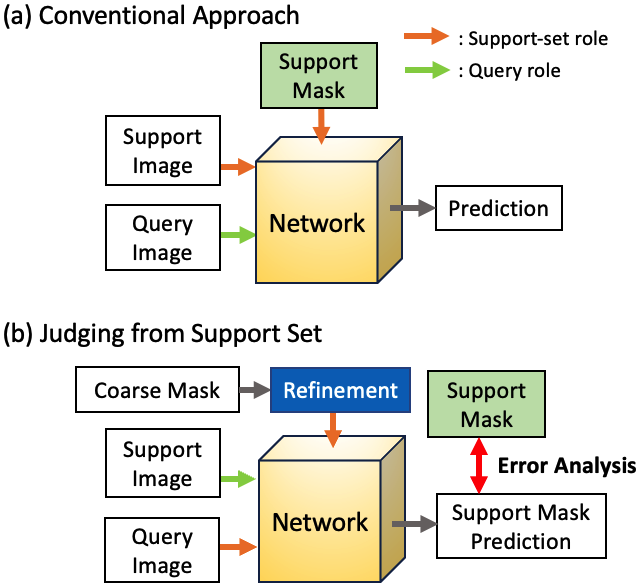}
 \caption{(a) The goal of conventional few-shot segmentation is to find the target object in the query image when a support set is given. (b) JFS approach. When there are a coarse mask and a refined mask, we can use the FSS network and support-set to prove that the refinement process is successful. Query image, segmentation masks~(Coarse mask prediction and refined prediction), and support-set used for deciding the success of refinement segmentation. Note: In this figure, the error analysis for a refined mask is only visualized.}
 \label{fig:comparision_fss}

 \vspace{-1em}

\end{figure}

\begin{figure*}[!h]
 \centering 
 \includegraphics[width=0.65\textwidth]{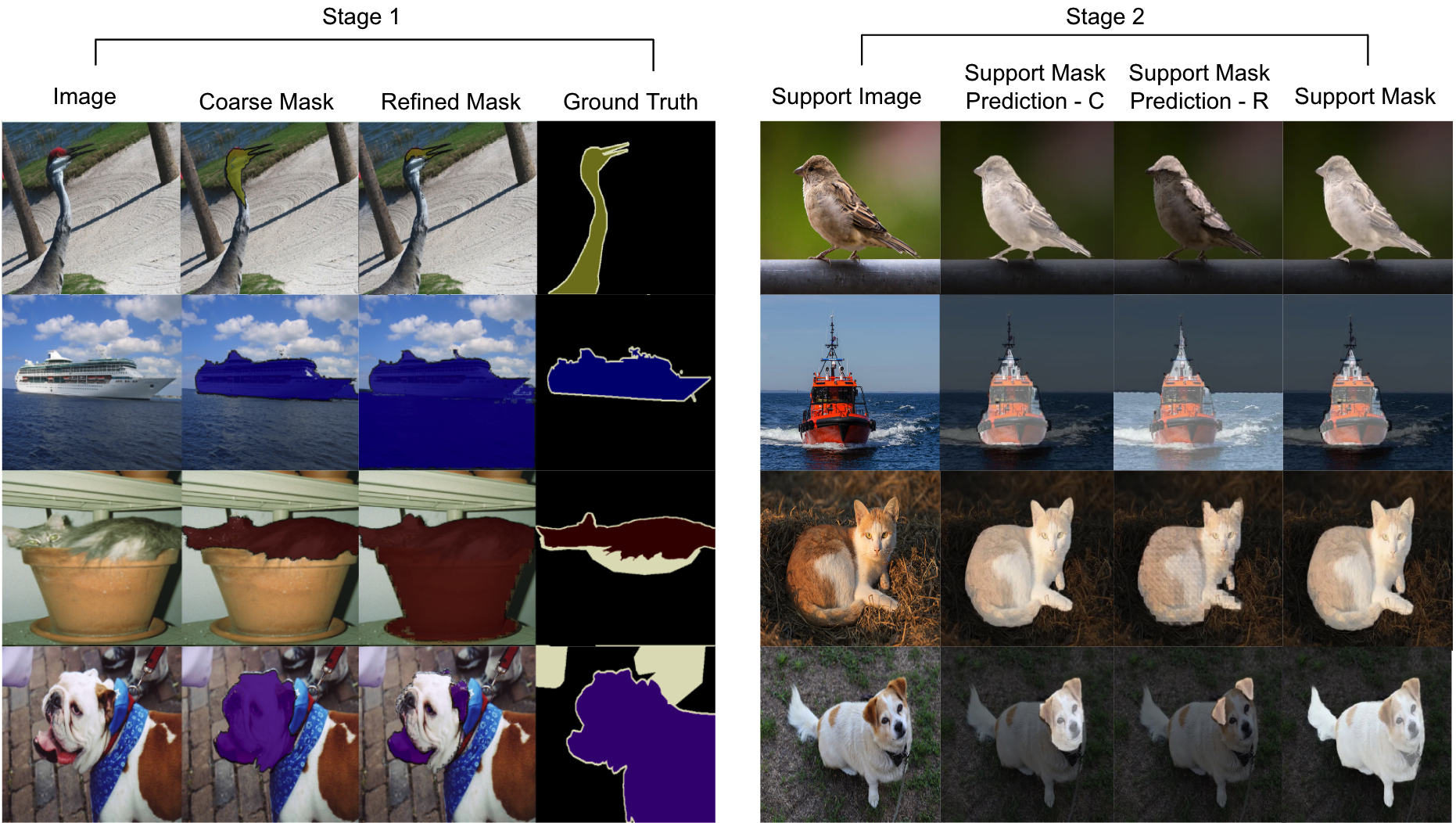}
 \caption{Segmentation refinement is a critical stage that aims to improve the initial, coarse masks generated by segmentation algorithms. These refined masks are supposed to capture better details and contours of the target objects. In the segmentation refinement process, as shown in the Stage 1 result on the left, problems can occur when the refined mask results in post-processing failures. The existing segmentation refinement has major limitations in its inability to determine whether refinement has failed because of non-existing ground truth information. We propose Judging from Support Set~(JFS), which can determine whether the refined mask has better results than the previous coarse mask or not. In Stage 2, Support mask prediction-C~($SMP^c$) is the result obtained using a coarse mask, and support mask prediction-R~($SMP^r)$ is the result obtained using a refined mask. The cases in Stage 2 show when the support mask prediction from the coarse mask performs better than the refined mask. $SMP^c$ closely matches the support mask. JFS can identify these cases by comparing the performance of coarse mask and refined mask in Stage 1 using FSS network. Therefore, we can identify whether segmentation refinement is successful or failed.}
 \label{fig:teaser}
\end{figure*}

\section{Introduction}
\label{sec:intro}

Semantic segmentation, which involves labeling each pixel of an image with a semantic category, is widely used in numerous fields, including medical imaging~\cite{yao2023medical}, semiconductor manufacturing~\cite{HUANG20151semiconductor}, and autonomous driving~\cite{autonomous}. Recent advancements in Deep Convolutional Neural Networks~(DCNNs) and the availability of extensive pixel-level annotated data have significantly enhanced the effectiveness of semantic segmentation techniques~\cite{Deeplab}. However, due to the complexity of directly generating high-quality masks, some prior research has focused on improving coarse masks generated by initial segmentation models. Previous works \cite{kirillov2020pointrend,ke2021masktransfiner} involve improving segmentation models with a custom module designed for mask correction. Other segmentation refinement studies~\cite{yuan2020segfix,SEPL} adopt model-agnostic approaches, using only an image and its coarse segmentation as input for refinement. Likewise, segmentation refinement continues to develop through various methods. 

Indeed, segmentation refinement can improve most initial segmentations; however,  it is important to note that segmentation refinement is not guaranteed to be successful~(Fig.~\ref{fig:teaser}) and the best results can only be expected when selecting an appropriate refinement model that performs well on a specific task. This makes determining the success and failure of segmentation refinement important. If the success and failure of segmentation refinement can be monitored, it would help to identify the most effective refinement techniques and achieve the best semantic segmentation. However, in contrast to the rapid developments of segmentation refinement techniques, we point out that the method for identifying whether the refinement is successful has been understudied despite its importance and usefulness. To address this research gap, we propose a new method that can assess the success of segmentation refinement using a few-shot segmentation~(FSS) network.


FSS is a class of methods that aims to address the shortcomings of supervised learning segmentation models. Specifically, FSS segments the target object in a query image using a few support image-segmentation pairs as prompts to reduce the cost and effort of curating new sets of data for retraining segmentation models from scratch. We found that FSS can also be applied to evaluate segmentation refinement methods. Since the support set prompting the FSS model contains ground-truth segmentation already, we can use them as queries instead and prompt the FSS model with the refined mask from the segmentation refinement modules. By comparing and evaluating the output of the FSS model prompted by the refined mask and the original coarse mask, we can determine whether the refined mask is truly an improvement over the coarse mask.  Based on this idea, we proposed a new method that can determine whether the segmentation refinement process is successful by using the support set and query images inversely from the conventional FSS approach (Fig.~\ref{fig:comparision_fss}). This study contributes to the field of segmentation refinement by offering a novel way to analyze the quality of refinement masks.

To the best of our knowledge, we are the first to introduce a method of using the FSS model to analyze the segmentation refinement process. Our method is named JFS~(\textbf{J}udging \textbf{F}rom \textbf{S}upport-Set) in the sense that it uses information from the support set in the FSS to determine the success of the refinement. Our proposed method can determine whether segmentation refinement was successful or failed. We summarize our key contributions
as follows:

\begin{itemize}
    \item We proposed JFS, which is able to leverage an off-the-shelf FSS network to evaluate segmentation refinement. We believe this is the first method to determine the success of segmentation refinement methods.
    \item We verify the effectiveness of our proposed method by combining JFS with the SAM Enhanced pseudo-label~(SEPL) pipeline on worst and best cases in PASCAL~\cite{pascal}.    
\end{itemize}

\section{Related work}

\subsection{Segmentation Refinement}

Segmentation refinement aims to improve the quality of masks generated by existing segmentation models. There are two main approaches to segmentation refinement: model-specific methods and model-agnostic methods. Model-specific methods involve designing algorithms that are tailored to a specific type of model.  For instance, PointRend~\cite{kirillov2020pointrend} and MaskTransfiner~\cite{ke2021masktransfiner}  are examples of model-specific methods.  PointRend~\cite{kirillov2020pointrend} conducts point-based segmentation predictions at dynamically chosen locations using an iterative subdivision algorithm. MaskTransfiner~\cite{ke2021masktransfiner}  constructs a quadtree from the sparse and incoherent regions within the RoI pyramid, and subsequently refines all tree nodes collectively using a refinement transformer that incorporates quadtree attention. Unlike model-specific methods, model-agnostic methods are more versatile in that they can refine masks from various models. For example, SegFix~\cite{yuan2020segfix} utilized boundary loss and direction loss separately on the predicted boundary map and direction map, respectively. BPR~\cite{refine4} extracts and enhances a series of boundary patches along the predicted instance boundaries using a powerful refinement network. SEPL~\cite{SEPL} employs CAM pseudo-labels as indicators to identify and merge SAM masks, producing high-quality pseudo-labels that are aware of both class and object. Segmentation refinement techniques have been widely used to improve coarse masks. However, no method has been developed that can identify the effectiveness of post-processing in improving initial segmentation. This study aims to address this research gap by proposing a new method, JFS.


\subsection{Few-shot Segmentation}
Few-shot segmentation~(FSS) aims to accurately segment a target object in a query image using a limited number of annotated examples called support-set. FSS was introduced by Shaban et al.~\cite{OSLSM}. Since then, many different methods have been attempted to improve FSS performance. PPNet~\cite{PPNet} and PMM~\cite{PMM} leveraged detailed object features and multiple prototypes for accuracy. DAN~\cite{DAN} focused on improving support and query image correspondence. HSNet~\cite{HSNet} proposed a 4D convolution for feature analysis, and VAT~\cite{VAT2} proposed a transformer-based network to model dense semantic correspondence. Various studies are continuously being conducted to improve FSS accuracy, but to our knowledge, there is no previous research using FSS to determine whether segment refinement is successful.

\section{Methodology}

\begin{figure}[!ht]
 \centering 
 \includegraphics[width=17em]{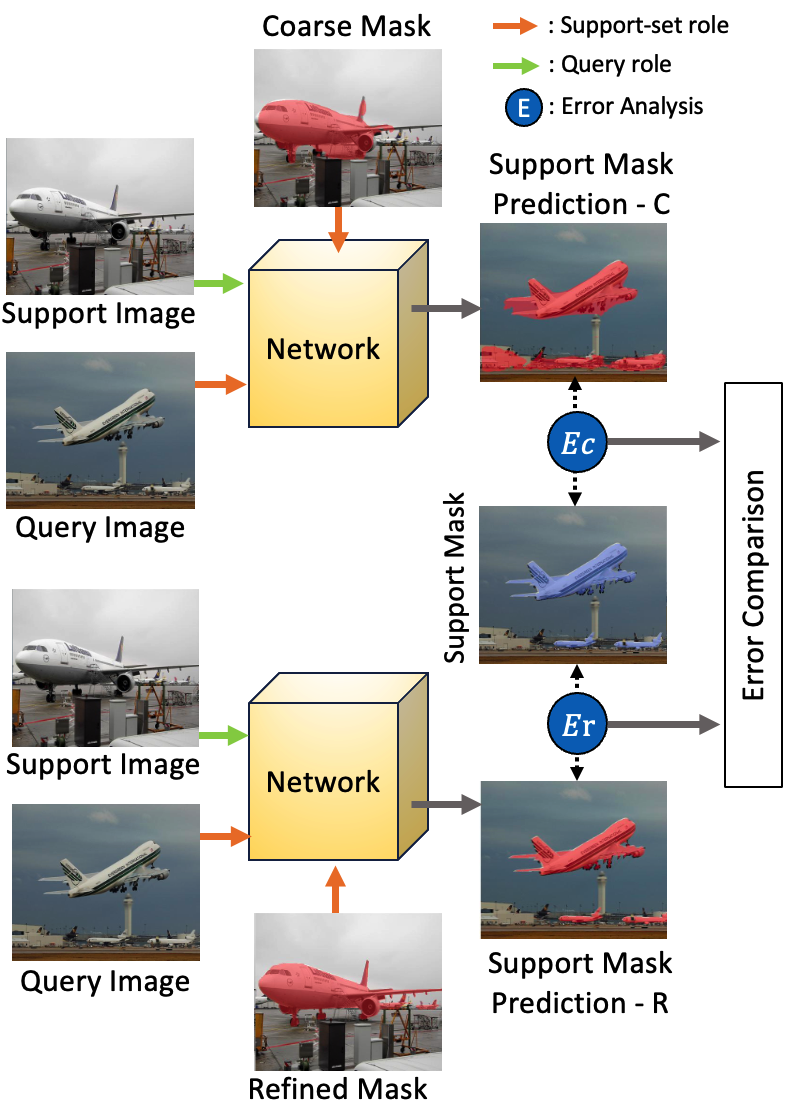}
 \caption{Overview of JFS. JFS uses support mask $M^s$ as ground truth to determine which mask prediction is closer to the ground truth. If the IoU value of $E_c$ is greater than $E_r$, segmentation refinement has failed. Conversely, if the IoU value of $E_r$ is greater than $E_c$, segmentation refinement has succeeded. Note: The terms Query role and Support-set role refer to their respective roles in Few-Shot Segmentation.}
 \label{fig:overview}
\end{figure}

The overall process of JFS is shown in Fig.~\ref{fig:overview}. First, we locate the target object using a segmentation model to obtain a coarse mask prediction. At this stage, the segmentation mask is often a coarse one. This is followed by the SAM Enhanced Pseduo-Labels~(SEPL)~\cite{SEPL} refinement process. During the refinement, the use of a robust Segmentation model, SAM~\cite{SAM}, allows for the acquisition of a more accurate segmentation mask of the object. However, whether the refined result is more accurate than the original prediction is questionable. The final step of our pipeline aims to answer this question. Specifically, we propose utilizing the FSS model to determine whether the refined segmentation is successful. The proposed framework is divided into two stages: (1) Coarse Mask Generation and Refinement and (2) Judging from Support set. This initial stage focuses on generating a preliminary segmentation mask, termed the `coarse mask'. Subsequently, this mask undergoes a refinement process to enhance its accuracy and detail, resulting in a `refined mask'. The second stage involves evaluating the refined masks by using them as support sets for FSS model. In this stage, another set of images along with their ground-truth segmentation is required for the evaluation, and we choose the original support set for the FSS model. These stages will be elaborated upon in the following discussion.

\subsection{Stage 1: Coarse Mask Generation and Refinement}


We utilized CAM~\cite{li2023transcam} to generate a coarse mask and the generated coarse mask is refined using SEPL~\cite{SEPL}. Coarse masks generated by CAM recognize distinct, class-specific regions but often miss precise object boundaries. We do not expect an accurate coarse mask because the purpose of the proposed method, JFS, is to determine whether segment refinement is successful or not. To get a refined mask, we utilized SEPL which improves the quality of coarse masks by using a strong segmentation tool, the Segment Anything Model~(SAM)~\cite{SAM}. Because SAM can accurately segment most parts or objects regardless of their class, SEPL can produce a high-quality segmentation mask. For one image, SEPL takes input coarse mask $[M^c_1, \cdots, M^c_K]$ and SAM masks $[S_0,  \cdots , S_L]$ and outputs refined masks $[M^r_1, \cdots, M^r_K]$. This process involves two steps: mask assignment and mask selection. In the mask assignment stage, SEPL calculates the intersection of each SAM mask $S_l$ with the coarse mask $M^c_k$ for each class $k$ from $1$ to $k$. Each mask $S_l$ is assigned to the class with which it overlaps the most, while masks with no overlap are discarded. 

Note that although SEPL~\cite{SEPL} generally improves segmentation quality, it has drawbacks from three sources: when a coarse mask fails to activate a target object, when the SAM fails to detect a target object, and the algorithm itself tends to have a large segment. These fail to improve segmentation quality from coarse masks and SEPL itself does not provide any information if refined masks have a worse quality mask than coarse masks.

\subsection{Stage 2: Judging from Support Set}


In this stage, we determine whether the refinement process is successful or not using the Algorithm~\ref{alg:two}~(Fig.~\ref{fig:overview}). With the two masks generated from stage 1, the coarse mask $M^c$ and the refined mask $M^r$, the query image $I^q$~(Original image used to create the coarse mask), the support image $I^s$ and the support mask $M^s$ are input to JFS. We avoid using the same image from PASCAL~\cite{pascal} for query and support images. Therefore, the query images are from the PASCAL train set, and the support image $I^s$ was taken from the PASCAL validation set. The support image should contain the same class object in the query image. All these inputs go to SegGPT~\cite{wang2023seggpt} which has shown a strong performance in the FSS task to get accurate support mask prediction~($SMP$). We get $SMP^c$ and $SMP^r$ from $M^c$ and $M^r$, respectively. We can obtain the IoU value by comparing the results of $SMP$ and $M^s$ using Equation~\ref{eq:1}. The higher the IoU value, the more similar it is to $M^s$. When $E_c$ is greater than $E_r$, it indicates that the segmentation refinement has failed. In contrast, if the value of $E_r$ is greater than $E_c$, it suggests that the refinement was successful.
\begin{align} \label{eq:1}
 E_c = \frac{SMP^c \cap M^s}{SMP^c \cup M^s}, \quad E_r = \frac{SMP^r \cap M^s}{SMP^r \cup M^s}
\end{align}

Note that other few-shot segmentation models can be utilized for this stage not only SegGPT.


\begin{algorithm}
\caption{JFS: Judging from Support Set}
\label{alg:two}
\textbf{Input}: Query Image $I^q$, Coarse Mask $M^c$, Refined Mask $M^r$, Support Image $I^s$, Support Mask $M^s$ 

\textbf{Output}: $E_c$ and $E_r$ scores 
\begin{algorithmic}[1]
\STATE Generate two sets of data {A and B}. $A = \{I^q, M^c, I^s \}$, $B = \{I^q, M^r, I^s \}$ 
\STATE Both $I^q, M^c$ and $I^q, M^r$ behave as Support-set role and $I^s$ becomes a query image in FSS 
\STATE Get Support Mask Prediction $SMP^c$ from A and $SMP^r$ from B using FSS network~(SegGPT) 
\STATE Calculate $E_c$ and $E_r$ comparing $SMP^c$, $SMP^r$ with $M^s$ respectively using Equation~\ref{eq:1}
\end{algorithmic}
\end{algorithm}






\subsection{Experiment}
\noindent\textbf{Design:} To evaluate JFS's ability to judge refined segmentations, we designed the following experiment. We use JFS to determine whether a refined mask improves over a coarse mask. If so, the refined mask is kept; otherwise, the coarse mask is retained. This process should result in a set of segment masks of higher quality by combining the best among the coarse and refined cases. We then measure the IOU of this new set against the original coarse and refined masks to assess the quality improvement achieved by JFS's judgment. We exclude samples where both the coarse mask and refined mask have an IoU of 0.


\noindent\textbf{Datasets:} For quantatative analysis, we evaluated the JFS method on the PASCAL~\cite{pascal} dataset. We generate coarse $M^c$ and refined mask $M^r$ for images $I^q$ in the PASCAL training set; then generate few-shot segmentation prediction $SMP^c$ and $SMP^r$ for the validation set images $I^s$ using Algorithm~\ref{alg:two}~. We perform analysis on several groups of instances: 1) Randomly select 400 images, 20 for each of the 20 classes; 2) Top and Bottom 50 improved SEPL outputs; 3) Top 20 improved SEPL outputs, and 4) Bottom 20 improved SEPL outputs. In addition, we use gathered license-free images per class for support-set from Wikimedia Commons for qualitative results.

\noindent\textbf{Evaluation metrics:}
We compared support mask prediction performance with GT support mask using Intersection-over-Union~(IoU) which is widely used in segmentation tasks~\cite{VAT2,hmmasking}. 
IoU calculates the overlap between the predicted region~(mask) and the ground truth region~(mask) of an object in an image, providing a measure of how accurately the predicted mask aligns with the ground truth mask.
\begin{equation}
\text{IoU} = \frac{|A \cap B|}{|A \cup B|}
\end{equation}
\( A \) represents the ground truth mask. \( B \) represents the prediction mask. \( |A \cap B| \) is the intersection area between the ground truth and prediction masks. \( |A \cup B| \) is the area of union of the ground truth and prediction masks. We calculate mean IoU, $mIoU =  \frac{1}{n}\sum_{1}^{n}IoU$ where $n$ is the number of test cases.

\noindent\textbf{Implementation Details:}  Coarse masks were created using TransCAM~\cite{li2023transcam}. We use SEPL~\cite{SEPL} method for refinement. All SAM~\cite{SAM} masks used in SEPL~\cite{SEPL} were generated using the pre-trained VIT-H model shared on the SAM official site. SegGPT was used for the FSS model, and the pre-trained model~(VIT large) was used without fine-tuning. The support images are drawn from the PASCAL validation set and are chosen to contain objects of the same class as the query image. We ensure that the support image and the query image are different images to prevent any data leakage.



\section{Result and Discussion}

\begin{table}[h!]
\begin{center}
\begin{tabular}{|c|c|c|c|c|}
\hline
Data & mIoU & \makecell{mIoU \\ (SEPL)}  & \makecell{mIoU \\ (JFS)} & \makecell{Success \\ Rate (\%)} \\ \hline
Random 400 & 59.8 & 64.5 & \textbf{65.2} & 52.1 \\ \hline
Top 50 + Low 50 & 48.6 & 62.2 & \textbf{67.5} & 62.1 \\ \hline
Top 20 & 29.8 & 88.6 & \textbf{89.6} & 68.4 \\ \hline
Low 20 & 65.4 & 33.2 & \textbf{50.5} & 60.0 \\ \hline
\end{tabular}
\end{center}
\caption{Comparison of mIoU of the coarse mask, the SEPL-refined mask, and JFS-selected segmentations, along with the success rate of JFS across different data selections. The success rate is measured as the percentage of correct decisions out of total samples. "Top 50/20" refers to the 50/20 samples with the greatest improvement in the refinement process, while "Low 50/20" denotes the 50/20 samples with the least improvement. The best results are highlighted in bold.}
\label{table2}
\end{table}


\noindent\textbf{Quantitative results:} 
Table~\ref{table2} compares the mIoU for coarse, refine and JFS picked segmentations, and shows the success rate of JFS across different experiment groups. The results show that the JFS picked segmentations consistently outperform the SEPL refinement in all four groups. The highest improvement by JFS was observed in the low 20 samples (where SEPL struggles), where JFS improved mIoU to 50.5 from SEPL's 33.2. In this group, both methods showed lower mIoU than the original coarse masks, but JFS improves on SEPL outputs by detecting failure cases in SEPL outputs. Furthermore, the success rate, which measures the effectiveness of JFS, is higher where mIoU is significantly improved by SEPL.
\noindent\textbf{Qualitative results:} Fig.~\ref{fig:teaser} shows that JFS can determine failure and success cases in the segmentation refinement process. Therefore, if the segmentation refinement fails, we can return to the existing coarse mask and consider other segmentation refinement methods. For this visualization, we used images publicly available on Wikimedia Commons for support-set. Query images are from PASCAL. We selected images where only one object was visible for each class. The support mask was obtained using the SAM segmentation method. 

\noindent\textbf{How JFS works?:} Most FSS networks~\cite{OSLSM,HSNet,VAT2,hmmasking,moon2023msi,moon2024fccfullyconnectedcorrelation} rely on high quality support set information to find target objects in the query image~(Fig.~\ref{fig:comparision_fss}). Therefore, FSS performance deteriorates when the support mask information is inaccurate~\cite{hmmasking,moon2023msi}. If a refined mask accurately segments the target object, it provides more detailed texture information of the target to FSS model. Hence, we designed JFS. 
We found from our qualitative analysis (Fig.~\ref{fig:teaser}) that the JFS method was particularly effective in the following cases~:

\textbf{(a)} When refined masks accurately segment the target object boundaries, support mask predictions will perform better at distinguishing the target object from the background.

\textbf{(b)} When the difference between the coarse and refined mask is clear, the presence of non-target objects in the refined mask confuses the FSS model, significantly dropping support mask prediction accuracy. Thus, it becomes evident whether the refined mask has failed or succeeded. Table~\ref{table2} shows the JFS success rate rises with significant mIoU improvement, indicating JFS's higher effectiveness when the refinement module makes greater improvement.
 
\textbf{(c)} When the query image and support-set have the same class object in the background. If the coarse mask or refined mask segments the target object and an object in the background, FSS models recognize this information as target information and attempt to segment it together if a similar object exists in the support image. 

\vspace{0.5em}




\begin{figure}[!htb]
 \centering 
 \includegraphics[width=17em]{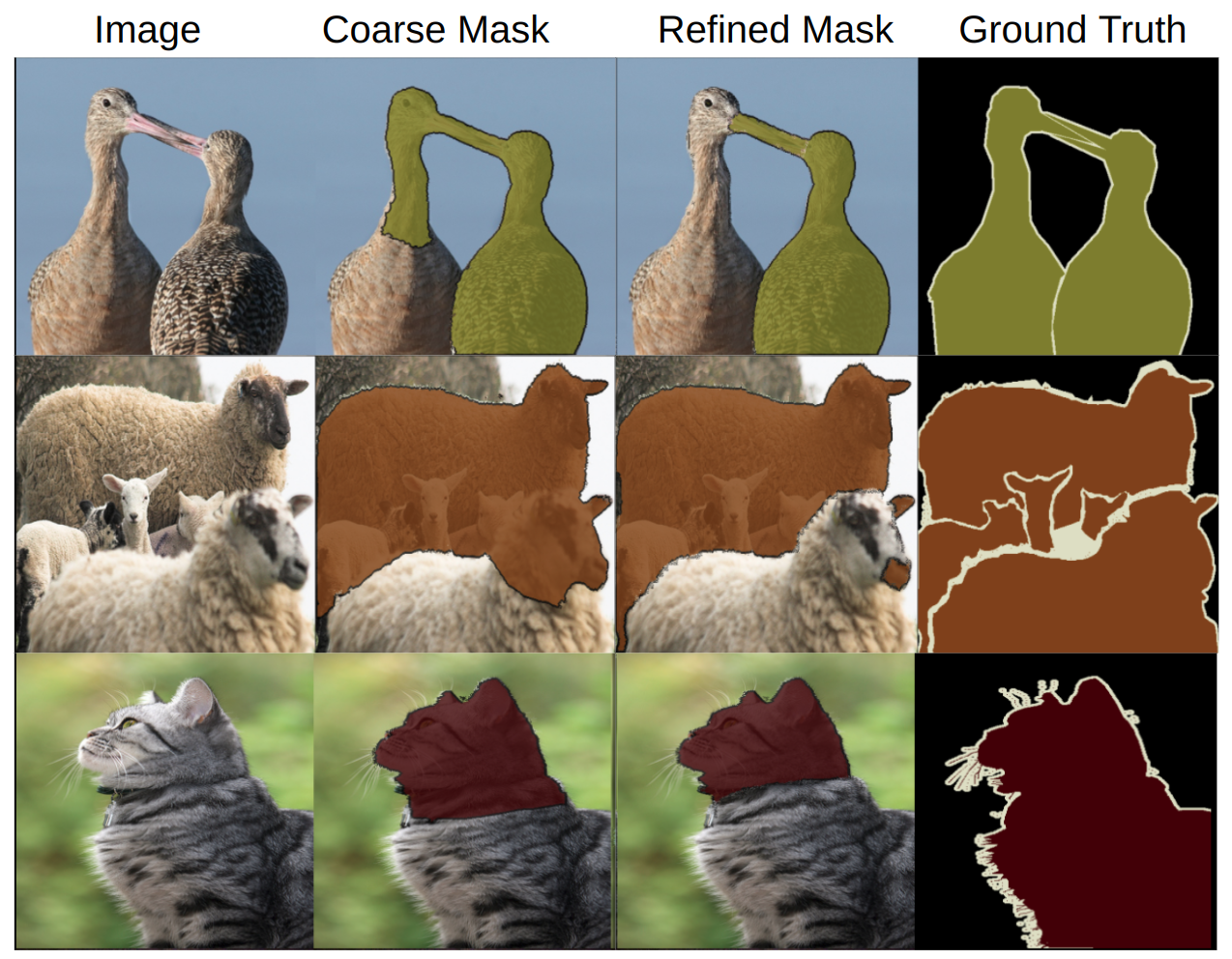}
 \caption{Failure cases. When the refined mask and coarse mask cover the same key features on the same target object, support mask prediction yields similar results. This is because FSS models are trained to find a target even with small key features from support-set. }
 \label{fig:failure}
\end{figure}

\vspace{-2em}

\section{Conclusion}

We proposed JFS, a novel method for assessing the effectiveness of the segmentation refinement process. JFS can distinguish between successful and failed cases in SEPL segmentation refinement. The identification of successful refinements can be applied in various scenarios, leading to improved segmentation quality. Furthermore, JFS introduces a fresh perspective in the few-shot segmentation and segmentation refinement domains by repurposing FSS for a unique application. However, we recognize that JFS underperforms in certain instances, indicating the need for further development~(Fig.~\ref{fig:failure}). Future research could achieve better results by fine-tuning FSS models or adapting them specifically for the JFS framework. 


\section{Acknowledgment}
This research used resources of the National Energy Research Scientific Computing Center, a DOE Office of Science User Facility supported by the Office of Science of the U.S. Department of Energy under Contract No. DE-AC02-05CH11231 using NERSC award NERSC DDR-ERCAP0033558.


\bibliographystyle{IEEEbib}
\bibliography{refs}

\begin{thebibliography}{10}

\bibitem{yao2023medical}
Wenjian Yao, Jiajun Bai, Wei Liao, Yuheng Chen, Mengjuan Liu, and Yao Xie,
\newblock ``From cnn to transformer: A review of medical image segmentation models,'' 2023.

\bibitem{HUANG20151semiconductor}
Szu-Hao Huang and Ying-Cheng Pan,
\newblock ``Automated visual inspection in the semiconductor industry: A survey,''
\newblock {\em Computers in Industry}, vol. 66, pp. 1--10, 2015.

\bibitem{autonomous}
Khan Muhammad, Tanveer Hussain, Hayat Ullah, Javier~Del Ser, Mahdi Rezaei, Neeraj Kumar, Mohammad Hijji, Paolo Bellavista, and Victor Hugo~C. de~Albuquerque,
\newblock ``Vision-based semantic segmentation in scene understanding for autonomous driving: Recent achievements, challenges, and outlooks,''
\newblock {\em IEEE Transactions on Intelligent Transportation Systems}, vol. 23, no. 12, pp. 22694--22715, 2022.

\bibitem{Deeplab}
Liang-Chieh Chen, George Papandreou, Iasonas Kokkinos, Kevin Murphy, and Alan~L. Yuille,
\newblock ``{DeepLab: Semantic Image Segmentation with Deep Convolutional Nets, Atrous Convolution, and Fully Connected CRFs},''
\newblock {\em IEEE Transactions on Pattern Analysis and Machine Intelligence (TPAMI)}, vol. 40, no. 4, pp. 834--848, 2018.

\bibitem{kirillov2020pointrend}
Alexander Kirillov, Yuxin Wu, Kaiming He, and Ross Girshick,
\newblock ``Pointrend: Image segmentation as rendering,'' 2020.

\bibitem{ke2021masktransfiner}
Lei Ke, Martin Danelljan, Xia Li, Yu-Wing Tai, Chi-Keung Tang, and Fisher Yu,
\newblock ``Mask transfiner for high-quality instance segmentation,'' 2021.

\bibitem{yuan2020segfix}
Yuhui Yuan, Jingyi Xie, Xilin Chen, and Jingdong Wang,
\newblock ``Segfix: Model-agnostic boundary refinement for segmentation,'' 2020.

\bibitem{SEPL}
Tianle Chen, Zheda Mai, Ruiwen Li, and Wei lun Chao,
\newblock ``Segment anything model (sam) enhanced pseudo labels for weakly supervised semantic segmentation,'' 2023.

\bibitem{pascal}
Mark Everingham, Luc Van~Gool, Christopher Williams, John Winn, and Andrew Zisserman,
\newblock ``{The Pascal Visual Object Classes (VOC) challenge},''
\newblock {\em International Journal of Computer Vision (IJCV)}, vol. 88, pp. 303--338, 06 2010.

\bibitem{refine4}
Chufeng Tang, Hang Chen, Xiao Li, Jianmin Li, Zhaoxiang Zhang, and Xiaolin Hu,
\newblock ``Look closer to segment better: Boundary patch refinement for instance segmentation,'' 2021.

\bibitem{OSLSM}
Amirreza Shaban, Shray Bansal, Zhen Liu, Irfan Essa, and Byron Boots,
\newblock ``{One-Shot Learning for Semantic Segmentation},''
\newblock in {\em Proceedings of British Machine Vision Conference (BMVC)}, 2017, pp. 167.1--167.13.

\bibitem{PPNet}
Yongfei Liu, Xiangyi Zhang, Songyang Zhang, and Xuming He,
\newblock ``Part-aware prototype network for few-shot semantic segmentation,'' 2020.

\bibitem{PMM}
Boyu Yang, Chang Liu, Bohao Li, Jianbin Jiao, and Ye~Qixiang,
\newblock ``{Prototype Mixture Models for Few-shot Semantic Segmentation},''
\newblock in {\em Proceedings of European Conference on Computer Vision (ECCV)}, 2020.

\bibitem{DAN}
Haochen Wang, Xudong Zhang, Yutao Hu, Yandan Yang, Xianbin Cao, and Xiantong Zhen,
\newblock ``{Few-Shot Semantic Segmentation with Democratic Attention Networks},''
\newblock in {\em Proceedings of European Conference on Computer Vision (ECCV)}, 2020, pp. 730--746.

\bibitem{HSNet}
Juhong Min, Dahyun Kang, and Minsu Cho,
\newblock ``{Hypercorrelation Squeeze for Few-Shot Segmentation},''
\newblock in {\em Proceedings of International Conference on Computer Vision (ICCV)}, 2021, pp. 6941--6952.

\bibitem{VAT2}
Sunghwan Hong, Seokju Cho, Jisu Nam, Stephen Lin, and Seungryong Kim,
\newblock ``{Cost Aggregation with 4D Convolutional Swin Transformer for Few-Shot Segmentation},''
\newblock in {\em Proceedings of European Conference on Computer Vision (ECCV)}, 2022.

\bibitem{SAM}
Alexander Kirillov, Eric Mintun, Nikhila Ravi, Hanzi Mao, Chloe Rolland, Laura Gustafson, Tete Xiao, Spencer Whitehead, Alexander~C. Berg, Wan-Yen Lo, Piotr Dollár, and Ross Girshick,
\newblock ``Segment anything,'' 2023.

\bibitem{li2023transcam}
Ruiwen Li, Zheda Mai, Zhibo Zhang, Jongseong Jang, and Scott Sanner,
\newblock ``Transcam: Transformer attention-based cam refinement for weakly supervised semantic segmentation,''
\newblock {\em Journal of Visual Communication and Image Representation}, p. 103800, 2023.

\bibitem{wang2023seggpt}
Xinlong Wang, Xiaosong Zhang, Yue Cao, Wen Wang, Chunhua Shen, and Tiejun Huang,
\newblock ``Seggpt: Segmenting everything in context,'' 2023.

\bibitem{hmmasking}
Seonghyeon Moon, Samuel~S. Sohn, Honglu Zhou, Sejong Yoon, Vladimir Pavlovic, Muhammad~Haris Khan, and Mubbasir Kapadia,
\newblock ``{HM: Hybrid Masking for Few-Shot Segmentation},''
\newblock in {\em Proceedings of European Conference on Computer Vision (ECCV)}, 2022, pp. 506--523.

\bibitem{moon2023msi}
Seonghyeon Moon, Samuel~S. Sohn, Honglu Zhou, Sejong Yoon, Vladimir Pavlovic, Muhammad~Haris Khan, and Mubbasir Kapadia,
\newblock ``Msi: Maximize support-set information for few-shot segmentation,''
\newblock in {\em Proceedings of the IEEE/CVF International Conference on Computer Vision (ICCV)}, October 2023, pp. 19266--19276.

\bibitem{moon2024fccfullyconnectedcorrelation}
Seonghyeon Moon, Haein Kong, Muhammad~Haris Khan, and Yuewei Lin,
\newblock ``Fcc: Fully connected correlation for few-shot segmentation,'' 2024.

\end{thebibliography}

\end{document}